\documentclass{article}

 \usepackage[preprint]{neurips_2026}

\usepackage[utf8]{inputenc} 
\usepackage[T1]{fontenc}    
\usepackage{hyperref}       
\usepackage{url}            
\usepackage{booktabs}       
\usepackage{amsfonts}       
\usepackage{nicefrac}       
\usepackage{microtype}      
\usepackage{xcolor}         

\usepackage{graphicx}
\usepackage{subcaption}

\usepackage{multirow}  
\usepackage{amsmath}
\usepackage{amssymb}
\usepackage{mathtools}
\usepackage{amsthm}
\usepackage{xcolor}
\newcommand{\authorskip}{\hspace{4mm}}
\title{Enhancing Video Physical Consistency via Role-aware Joint Training and Modality-decoupled Denoising}

\author{%
  Guangting Zheng$^{1}$\thanks{Equal contribution to this work.}\authorskip Haojing Chen$^{2}$\footnotemark[1]\authorskip Hao Li$^{3}$\authorskip Jingtao Zhang$^{4}$\authorskip Zhen Yang$^{1}$\\ \textbf{Xiaosong Jia$^{3}$\authorskip Xue Yang$^{5}$\authorskip Shaofeng Zhang$^{1}$\thanks{Corresponding author.}\authorskip Yanyong Zhang$^{1}$\footnotemark[2]}\\
  $^{1}$University of Science and Technology of China\\
  $^{2}$University of Electronic Science and Technology of China\authorskip
  $^{3}$Fudan University\\
  $^{4}$Georgia Institute of Technology\authorskip
  $^{5}$ Shanghai Jiao Tong University
}

\begin{document}

\maketitle

\begin{abstract}
  While modern video diffusion models excel in visual fidelity, maintaining long-range physical consistency remains a formidable challenge. Conventional pixel-reconstruction objectives mainly focus on appearance details and often fail to capture the underlying dynamics of a scene. To mitigate this, recent efforts have integrated auxiliary modalities (e.g., optical flow) to introduce physics priors via joint training with video appearance. However, these methods have three main limitations: (1) they do not distinguish the different motion patterns of different entity types; (2) joint modeling of visual and auxiliary modalities can cause capacity conflicts and weaken the pretrained visual prior; and (3) auxiliary modalities may accumulate errors during inference. To address these issues, we propose \textbf{VPT}, a fine-tuning framework for improving physical consistency in video diffusion models. VPT introduces a role-aware signal that groups entities into agents, controlled objects, passive objects, and background, so that different physical roles can be modeled more clearly. We further propose a modality-decoupled denoising strategy, where the visual and auxiliary channels are assigned independent noise levels. Together with a loss-weight decay strategy, this design makes auxiliary modalities serve as soft constraints rather than strong dependencies, mitigating recursive prediction errors during inference. We also introduce cross-step auto-guidance to further strengthen physical dynamics.
  Experiments show that VPT improves physical consistency while preserving visual quality, achieving relative gains of 39.4\% in SA and 17.9\% in PC on VideoPhy benchmark over Wan2.1-T2V-1.3B, and consistent improvements on VideoPhy-2 benchmark. The project page is available at \url{https://tom-zgt.github.io/VPT}.
\end{abstract}

\section{Introduction}
\begin{figure}[t]
    \centering
    \begin{subfigure}[t]{\linewidth}
        \centering
        \includegraphics[width=\linewidth]{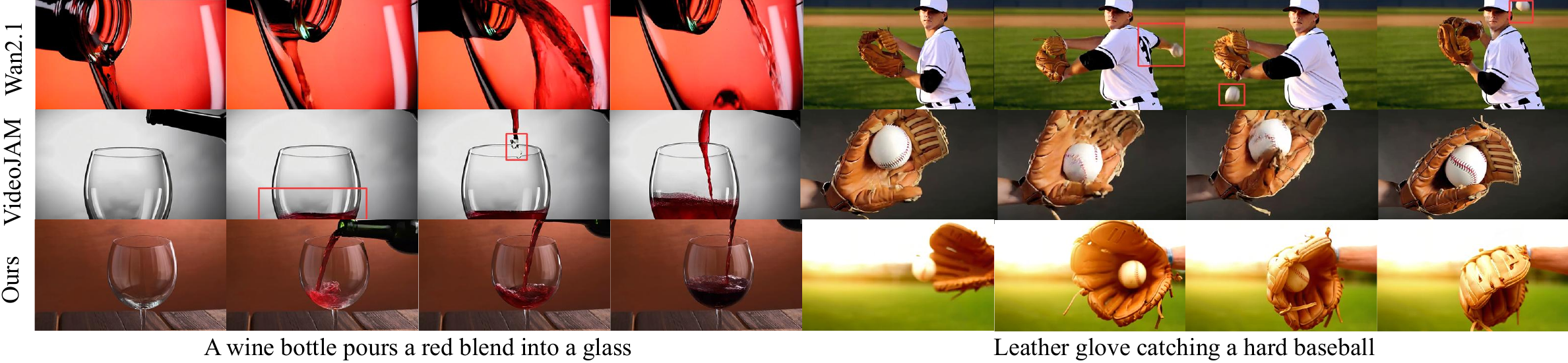}
        \caption{Text-to-Video samples generated by Wan2.1-T2V-1.3B, VideoJam, and our VPT, showing that VPT produces more physically consistent object interactions and motion.}
        \label{fig:intro_examples}
    \end{subfigure}

    \begin{subfigure}[t]{\linewidth}
        \centering
        \includegraphics[width=\linewidth]{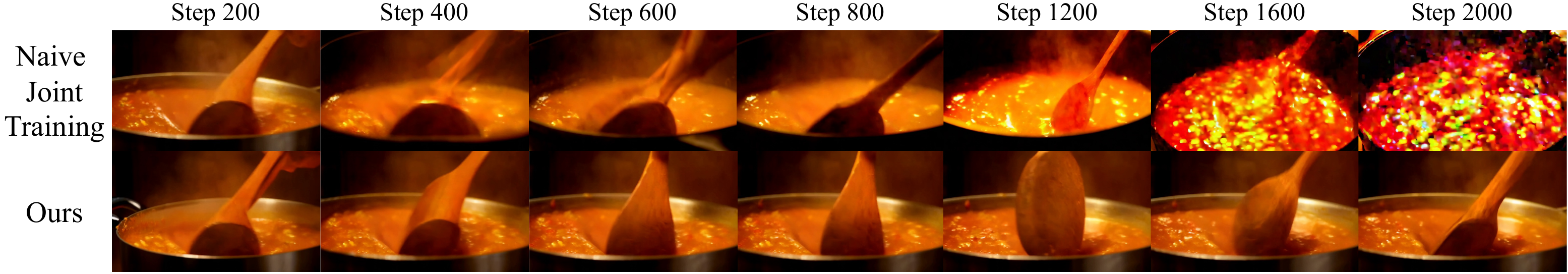}
        \caption{Comparison between naive joint appearance-motion training and VPT, where naive joint training shows clear visual collapse around step 1200 and further deteriorates afterwards, while VPT maintains stable visual quality.}
        \label{fig:motivations_c}
    \end{subfigure}

    \begin{subfigure}[t]{0.48\linewidth}
        \centering
        \includegraphics[width=\linewidth]{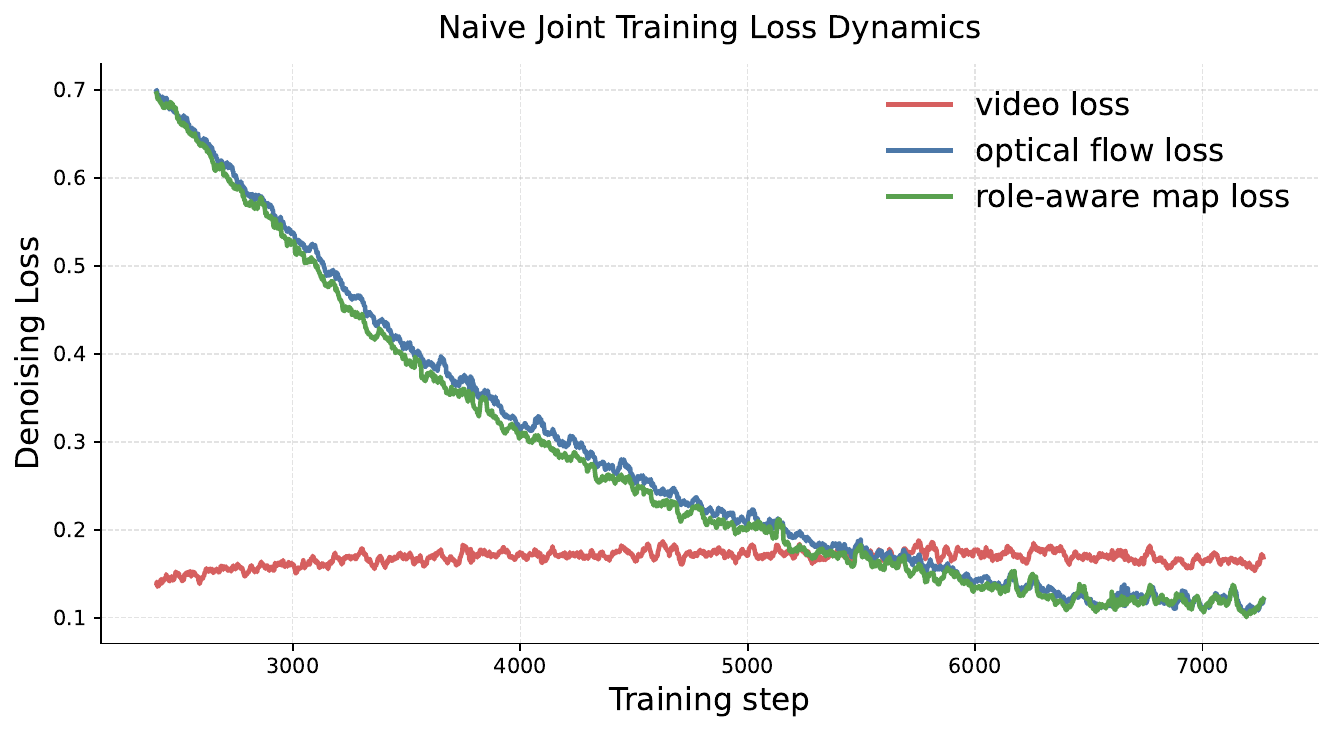}
        \caption{Loss dynamics of naive joint training, where decreasing auxiliary losses coincide with an increasing video denoising loss, indicating capacity conflict. }
        \label{fig:motivations_a}
    \end{subfigure}
    \hfill 
    \begin{subfigure}[t]{0.48\linewidth}
        \centering
        \includegraphics[width=\linewidth]{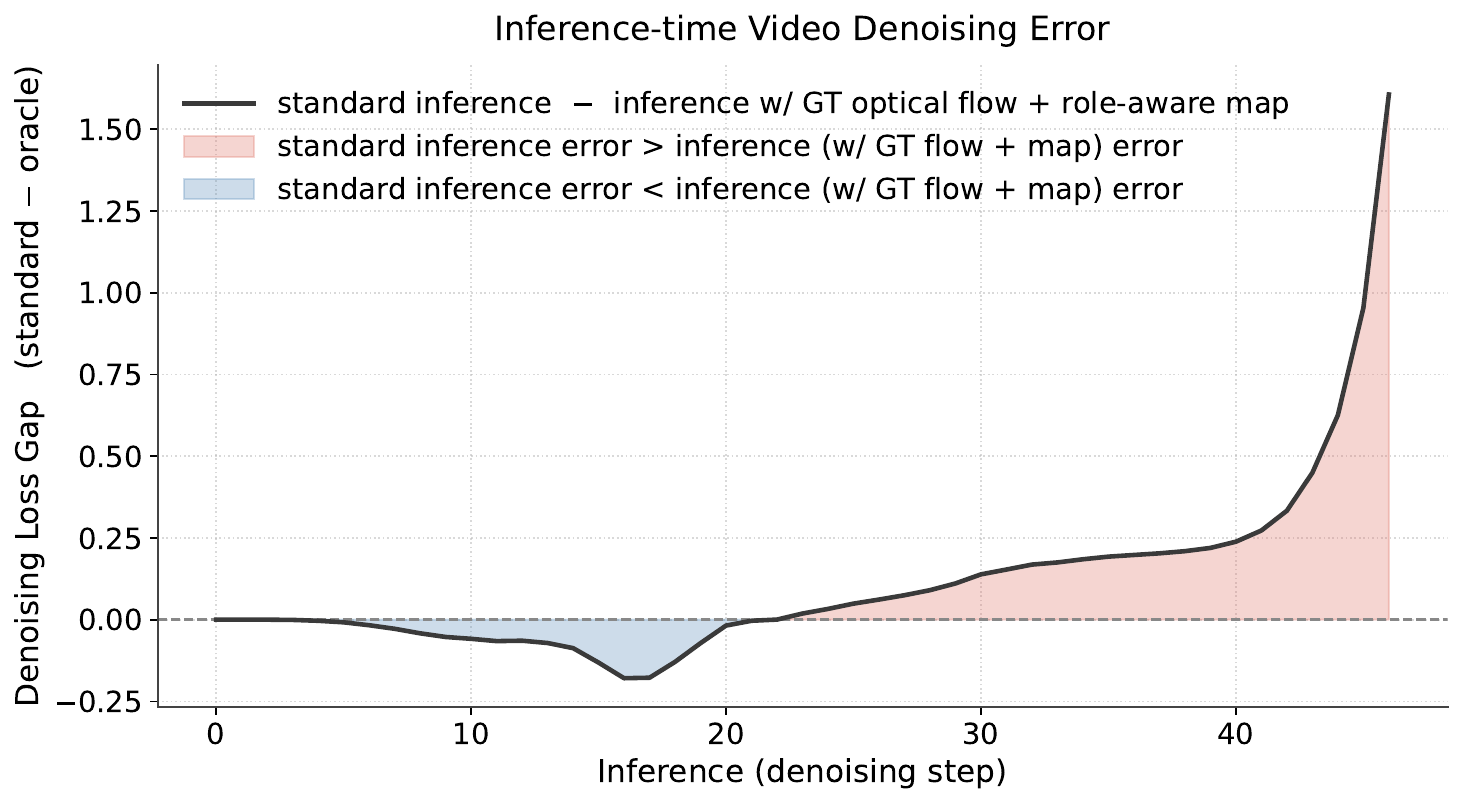}
        \caption{Inference-time denoising error caused by recursively predicted auxiliary modalities, illustrating error accumulation during sampling.}
        \label{fig:motivations_b}
    \end{subfigure}

    \caption{\textbf{Overview of qualitative examples and motivation experiments}.}
    \label{fig:combined_figure}
\end{figure}

Video diffusion models have achieved remarkable progress, demonstrating strong capability in generating high-fidelity videos with rich visual details~\cite{bartal2024lumiere,goku,nova,gen3,hunyuanvideo,opensora,infstar,wan2025wan,yang2024cogvideox}. Despite this, maintaining long-range physical consistency in complex dynamic scenes remains challenging~\cite{kang2024how,motamed2025generative,zhang2025videorepa}. For example, as shown in Figure \ref{fig:intro_examples}, when a wine bottle pours wine into a glass, the wine may follow an implausible trajectory, or its surface may deform unnaturally during pouring. A key reason is that standard pixel-reconstruction objectives primarily emphasize high-frequency appearance details, while providing limited supervision for underlying physical dynamics~\cite{chefer2025videojam}.

To mitigate these issues, recent methods~\cite{chefer2025videojam,chen20254dnex,wonder3d}, such as VideoJAM~\cite{chefer2025videojam}, introduce auxiliary modalities such as optical flow and jointly train them with video appearance. 
Although these modalities provide useful motion priors, existing approaches still face three key limitations. 
First, they usually model optical flow uniformly and ignore the heterogeneous dynamics of different entities. 
For example, an agent, a manipulated object, and a passive object often follow different motion patterns. 
As shown in Figure~\ref{fig:intro_examples}, the human initiates the action, the leather glove is controlled by the agent, and the baseball is passively affected. 
Second, joint modeling of appearance and auxiliary modalities may introduce capacity conflicts. As shown in the motivation experiment in Figure~\ref{fig:motivations_a}, when the model is optimized to fit both visual appearance and auxiliary modalities, their objectives can compete with each other. We observe that as the auxiliary modality denoising losses decrease, the video denoising loss begins to increase. Continuing joint training under such conflict can weaken the pretrained visual prior, leading to degraded visual quality. As shown in Figure~\ref{fig:motivations_c}, naive joint training progressively reduces video quality and may even cause generation collapse, whereas our method maintains stable and high-quality outputs. Third, existing methods are prone to severe inference-time error accumulation. This issue arises because the denoising process depends on the model's own auxiliary predictions, which are often less reliable than visual predictions. The visual backbone is typically pretrained on large-scale video data for extensive iterations, whereas auxiliary modalities receive much less training. As shown in the motivation experiment in Figure \ref{fig:motivations_b}, even minor errors in auxiliary predictions can be propagated across denoising steps and progressively amplified.

To address these limitations, we propose VPT, short for Video Physical-consistency Tuning. VPT is a fine-tuning framework designed to improve physical consistency while preserving the visual quality of pretrained video diffusion models. It consists of two key components.

First, we propose a role-aware joint representation. Instead of describing the entire scene solely with a dense motion field, VPT introduces a role-aware map that categorizes entities into four roles: agents, controlled objects, passive objects, and background. This map serves as a lightweight semantic anchor, introducing physical role information without substantial modeling overhead. It enables the model to distinguish different physical roles and better capture their interactions. 
As shown in the baseball-catching example in Figure~\ref{fig:intro_examples}, optical-flow-only methods such as VideoJAM provide dense motion cues but cannot distinguish the roles of different entities. 
The leather glove is agent-controlled and should actively catch the ball, while the baseball is passive and should change its motion upon contact. 
By encoding such role information, VPT better captures entity-specific dynamics and contact interactions than pure optical-flow supervision. 
With this lightweight design, VPT brings substantial improvements over the optical-flow-only variant on VideoPhy, demonstrating the importance of explicitly modeling entity-level physical roles.

Second, we propose modality-decoupled denoising with auxiliary loss annealing. 
Previous joint-training methods denoise video and auxiliary modalities at a shared timestep \(t\), creating a strong dependence between them.
In contrast, VPT assigns separate timesteps \((t_v, t_{f}, t_r)\) to video, optical flow, and role-aware signals, respectively. 
This decoupling allows auxiliary modalities to provide soft physical guidance without becoming mandatory targets, reducing their adverse impact when auxiliary predictions are inaccurate.
We further apply loss-weight decay to gradually reduce the influence of auxiliary losses so that the model can learn useful physical priors while preserving the pretrained visual prior. 

Since predicted auxiliary modalities can be noisy and inaccurate, directly using them as inference-time inner-guidance~\cite{wan2025wan} often provides limited benefits and may even degrade performance.
We instead introduce cross-step auto-guidance, which uses an intermediate model as a reference to amplify only the physical direction learned by the final model, while preserving its visual quality.

Experiments demonstrate that VPT effectively improves the physical consistency of pretrained video diffusion models, increasing SA/PC from 47.7/21.2 to 66.5/25.0 on VideoPhy~\cite{bansal2024videophy} and increasing SA/PC from 19.3/53.7 to 22.5/55.1 on VideoPhy2~\cite{bansal2025videophy2}, while maintaining competitive visual quality on VBench~\cite{huang2023vbench}. 
Overall, VPT provides an effective and lightweight fine-tuning framework for improving physical consistency in pretrained video diffusion models.

\section{Related Work}
\paragraph{Video Diffusion Models.} Diffusion-based video generation has evolved from early cascaded and U-Net-based architectures to scalable latent diffusion and transformer-based frameworks. Early works extend image diffusion models to videos through temporal layers, cascaded spatial-temporal generation, latent-space modeling, or motion modules~\cite{ho2022video, ho2022imagen, singer2022make, blattmann2023align, blattmann2023stable, guo2024animatediffanimatepersonalizedtexttoimage, bartal2024lumiere}. More recently, diffusion transformers~\cite{peebles2023scalable} and flow-matching models~\cite{lipman2022flow} have become dominant backbones for large-scale text-to-video generation. Representative models such as Latte~\cite{ma2024latte}, CogVideoX~\cite{yang2024cogvideox}, HunyuanVideo~\cite{hunyuanvideo}, and Wan~\cite{wan2025wan} demonstrate strong scalability, fidelity, and prompt alignment. Despite these advances, existing video diffusion models are still mainly optimized by reconstructing pixels or latent velocities~\cite{ho2020denoising,rombach2022high,lipman2022flow}, which provides limited supervision for long-range dynamics and physical interactions~\cite{chefer2025videojam,bansal2024videophy,meng2024worldsimulatorcraftingphysical,bansal2025videophy2}. As a result, generated videos may preserve high per-frame visual quality while still violating physical consistency in object motion, contact, deformation, and material interaction.

\paragraph{Video Physical Consistency.}
Recent benchmarks show that video generative models still struggle with physical commonsense and action-level rules~\cite{bansal2024videophy,meng2024worldsimulatorcraftingphysical,bansal2025videophy2}. Existing methods improve physical consistency by introducing structured priors beyond RGB reconstruction. Simulation-based methods, such as PhysGen~\cite{liu2024physgen} and MotionCraft~\cite{montanaro2024motioncraft}, use physical simulation or dynamics-aware motion fields, but are often limited by simulator assumptions and predefined scenarios. Non-simulation methods inject physical or world knowledge through reasoning, physics-aware data, or representation alignment, as in PhyT2V~\cite{xue2024phyt2v}, WISA~\cite{wang2025wisa}, and VideoREPA~\cite{zhang2025videorepa}. Related works also model auxiliary modalities, such as appearance-motion representations in VideoJAM~\cite{chefer2025videojam}, dynamic 3D/4D geometry in 4DNeX~\cite{chen20254dnex}, and multi-modal signals in OmniVDiff and UnityVideo~\cite{xi2025omnivdiff,huang2025unityvideo}. However, these methods mainly rely on dense motion, geometry, global representations, or predefined physical categories, and rarely explicitly model the role-dependent dynamics of different entities. In contrast, VPT introduces a lightweight role-aware signal that separates agents, controlled objects, passive objects, and background, enabling entity-level physical consistency modeling.

\section{Methodology}
\label{sec:method}
We propose VPT, a fine-tuning framework for enhancing physical consistency in video diffusion models. 
VPT consists of three key components. 
First, it introduces a role-aware joint representation that augments optical flow with entity-level physical roles. 
Second, it adopts modality-decoupled denoising with auxiliary loss annealing, which reduces optimization conflicts between visual generation and auxiliary prediction and mitigates inference-time error accumulation. 
Third, VPT introduces cross-step auto-guidance during inference, which improves motion dynamics.

\subsection{Preliminaries: Joint Appearance-Motion Training}
Modern video diffusion models commonly perform denoising in a latent space~\cite{wan2025wan,rombach2022high,blattmann2023stable}. 
Let \(z^v\) denote the video latent and \(z^f\) denote the optical-flow latent. 
Given a text condition \(y\), existing joint appearance-motion training methods concatenate visual and auxiliary latents and train the model to predict their joint velocity field under a shared diffusion timestep \(t\):
\begin{equation}
\mathcal{L}_{\mathrm{joint}}
=
\mathbb{E}_{t, z^v, z^f, y}
\left[
\left\|
u_{\theta}
\left(
[z^v_t \oplus z^f_t], y, t
\right)
-
[v^v_t \oplus v^f_t]
\right\|_2^2
\right],
\label{eq:joint_training}
\end{equation}
where \(u_{\theta}\) is the velocity prediction network, \(v^v_t\) and \(v^f_t\) are the target velocity fields for video and optical flow, respectively, and \(\oplus\) denotes channel-wise concatenation. 
In practice, this formulation requires expanding the input and output projection layers of the pretrained video diffusion model to accommodate additional auxiliary channels.
Although this synchronized joint-training paradigm provides explicit motion supervision, it tightly couples visual and auxiliary denoising throughout both training and inference. 
Such coupling can introduce optimization conflicts during fine-tuning and cause auxiliary prediction errors to propagate during sampling.

\subsection{Motivation: Limitations of Joint Appearance-Motion Training}
Despite its simplicity, the synchronized joint-training formulation in Eq.~\eqref{eq:joint_training} has three key limitations.
\paragraph{Heterogeneous dynamics.}
Optical flow describes pixel-level displacement but does not distinguish the physical roles of different entities. 
However, in interactive scenes, different entities often follow distinct motion patterns. 
For example, an agent that initiates an action, an object manipulated by the agent, and a passive object affected by external forces should not be modeled with the same motion prior. 
Treating all regions uniformly fails to capture such role-dependent dynamics. 
This motivates us to introduce a role-aware representation that explicitly encodes entity-level physical roles.
\paragraph{Capacity conflict.}
Jointly predicting visual and auxiliary modalities requires the pretrained video model to allocate part of its modeling capacity to newly introduced auxiliary channels. 
This can conflict with the original visual denoising objective and weaken the pretrained visual prior. 
As shown in Figure~\ref{fig:motivations_a}, during fine-tuning, the denoising losses of auxiliary modalities decrease, while the video denoising loss begins to increase. 
This indicates that optimizing auxiliary prediction can come at the cost of visual modeling quality.
Moreover, synchronized joint training enforces all modalities to be denoised at the same timestep \(t\), as shown in Eq.~\eqref{eq:joint_training}. 
Since optical flow and role-aware signals are highly correlated with video motion, the model may exploit them as shortcuts for video denoising. 
Such shortcut dependence makes the visual branch strongly conditioned on auxiliary channels during training. 
However, unlike the visual branch, which benefits from large-scale pretraining over massive video data, the auxiliary branches are learned only during limited fine-tuning. 
As a result, the model can easily become over-dependent on auxiliary modalities that are themselves not sufficiently well learned, leading to a mismatch between training and inference.
\paragraph{Inference-time error accumulation.}
During training, auxiliary latents are derived from ground-truth modalities and therefore provide relatively reliable supervision. 
During inference, however, auxiliary modalities must be predicted by the model itself and are recursively used in subsequent denoising steps. 
Since auxiliary prediction is much less mature than visual generation, its errors can be propagated and progressively amplified throughout sampling, as shown in Figure~\ref{fig:motivations_b}.
Therefore, directly relying on recursively predicted auxiliary modalities during inference leads to severe error accumulation and unstable motion generation.

These observations motivate the design of VPT. 
To address heterogeneous dynamics, we propose a role-aware joint representation. 
To reduce capacity conflict and mitigate inference-time auxiliary drift, we introduce modality-decoupled denoising with independent timesteps and auxiliary loss annealing. 
\begin{figure}
    \centering
    \includegraphics[width=\linewidth]{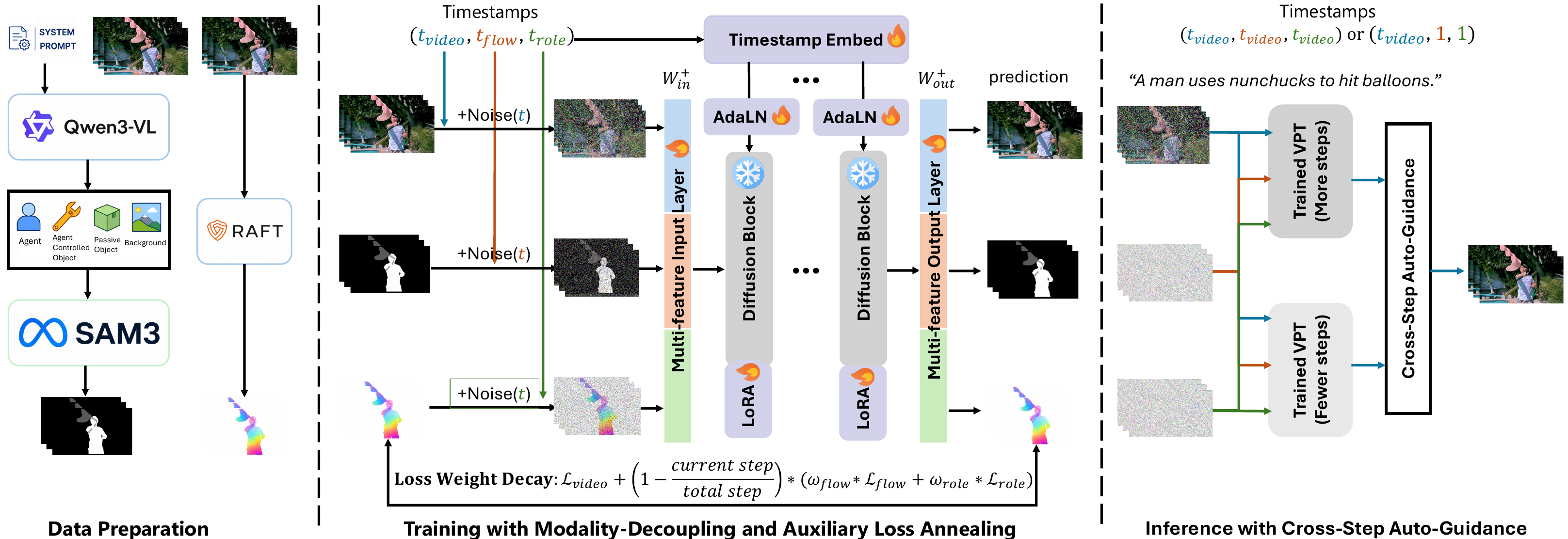}
    \caption{\textbf{Overview of VPT}. 
VPT consists of three stages: data preparation, modality-decoupled fine-tuning, and cross-step auto-guidance inference. 
First, optical flows and role-aware maps are extracted from training videos as auxiliary physical cues. 
Then, VPT finetunes the video diffusion model with modality-decoupled denoising and auxiliary loss annealing, decoupling visual and auxiliary denoising while gradually reducing auxiliary supervision.
Finally, cross-step auto-guidance uses an earlier intermediate model to guide the final model at inference time, improving motion dynamics.}
    \label{fig:method}
\end{figure}
\subsection{Role-Aware Joint Representation}

To model heterogeneous dynamics, VPT constructs a role-aware joint representation from a optical flow map and a role-aware map.

For each training video, we first extract the optical flow $\mathcal{F} \in \mathbb{R}^{T \times H \times W \times 3}$ using a pretrained RAFT model~\cite{teed2020raft}.
In parallel, we identify scene entities and assign each entity to one of four roles: agent $\mathcal{A}$, controlled object $\mathcal{O}_c$, passive object $\mathcal{O}_p$, and background $\mathcal{B}$. In implementation, we use Qwen3-VL~\cite{bai2025qwen3vltechnicalreport} for role assignment and SAM3~\cite{carion2026sam3segmentconcepts} to obtain temporally consistent masks. The role labels are converted into a dense role map $\mathcal{R} \in \mathbb{R}^{T \times H \times W \times 1}$.
We encode the four roles with fixed scalar values: $\{\mathcal{A}: 0,
\mathcal{O}_c: 85,
\mathcal{O}_p: 170,
\mathcal{B}: 255\}$.
The role map is then replicated along the channel dimension to obtain a three-channel representation compatible with the video VAE.

Both the optical flow $\mathcal{F}$ and the role map $\mathcal{R}$ are encoded by the original video VAE encoder:
\begin{equation*}
z^f = E_{\mathrm{VAE}}(\mathcal{F}),
\qquad
z^r = E_{\mathrm{VAE}}(\mathcal{R}),
\end{equation*}
where $z^f$ and $z^r$ denote the flow latent and the role latent, respectively. Using the original video VAE avoids introducing an additional feature space and keeps the auxiliary latents aligned with the pretrained visual latent distribution. We also empirically verify that the video VAE faithfully reconstructs optical flow and role-aware maps in Table \ref{tab:ablation} and Appendix \ref{app:aux_reconstruction}. The joint latent representation is formed by channel-wise concatenation $Z =[z^v \oplus z^f \oplus z^r]$. To improve robustness and support classifier-free guidance~\cite{ho2022classifierfree}, we randomly drop each auxiliary latent during training:
\begin{equation*}
\hat{z}_a =
\begin{cases}
\mathbf{0}, & \text{with probability } p, \\
z_a, & \text{with probability } 1-p,
\end{cases}
\qquad
a \in \{f, r\}.
\end{equation*}
The actual training input is therefore $\hat{Z} =
[z^v \oplus \hat{z}_f \oplus \hat{z}_r]$.

\subsection{Modality-Decoupled Denoising and Auxiliary Loss Annealing}

The synchronized objective in Eq.~\eqref{eq:joint_training} tightly couples visual and auxiliary denoising through a shared timestep, leading to a strong dependence between video generation and auxiliary prediction. 
To avoid such rigid dependence while still exploiting auxiliary physical priors, VPT decouples the denoising process across modalities and gradually anneals the auxiliary supervision during finetuning.

\paragraph{Modality-Decoupled Denoising.}

To reduce this rigid coupling, VPT introduces modality-decoupled denoising as a unified training strategy. 
Instead of using a shared timestep, we sample independent timesteps for the visual, optical-flow, and role-aware modalities, \(t_v \sim \mathcal{U}(0,1)\), \(t_f \sim \mathcal{U}(0,1)\), and \(t_r \sim \mathcal{U}(0,1)\). 
The noisy joint latent is constructed as $
\hat{Z}_{t_v,t_f,t_r}
=
[
z^v_{t_v}
\oplus
\hat{z}^f_{t_f}
\oplus
\hat{z}^r_{t_r}
],
$
where \(\hat{z}^f\) and \(\hat{z}^r\) denote the auxiliary latents after auxiliary dropout. 
The model takes all timesteps as conditions and predicts the velocity fields for all modalities:
\begin{equation*}
(\hat{v}^v, \hat{v}^f, \hat{v}^r)
=
u_{\theta}(\hat{Z}_{t_v,t_f,t_r}, y, t_v, t_f, t_r).
\end{equation*}
We compute modality-wise denoising losses as $
\mathcal{L}_v=\|\hat{v}^v-v^v\|_2^2,
\mathcal{L}_f=\|\hat{v}^f-v^f\|_2^2,
\mathcal{L}_r=\|\hat{v}^r-v^r\|_2^2
$. By decoupling \(t_v\), \(t_f\), and \(t_r\), the model is no longer required to reconstruct visual and auxiliary modalities at the same noise level. 
Thus, auxiliary signals provide structural guidance to the visual branch without imposing a strict synchronized prediction constraint. 
In other words, the auxiliary modalities are converted from hard denoising targets into soft physical constraints.

\paragraph{Auxiliary Loss Annealing.}

To further reduce the competition between visual generation and auxiliary prediction, we apply auxiliary loss annealing. 
At training iteration \(k\), the overall objective is
\begin{equation}
\mathcal{L}(k)
=
\mathcal{L}_v
+
\lambda(k)(\mathcal{L}_f+\mathcal{L}_r),
\qquad
\lambda(k)
=
\frac{\lambda_0}{2}
\left(
1+\cos\frac{\pi k}{K}
\right),
\end{equation}
where \(\lambda_0\) is the initial auxiliary weight and \(K\) is the total number of fine-tuning iterations. 
Early in training, the auxiliary losses encourage the model to acquire motion structure and role-aware physical bias. 
As training proceeds, their influence gradually decays, allowing the optimization to focus on the visual branch and better preserve the pretrained visual prior.

\subsection{Cross-step Auto-Guidance}

Auxiliary loss annealing produces a sequence of intermediate models at different finetuning steps, each reflecting a different degree of physical-prior absorption.
As finetuning proceeds, the model progressively internalizes the motion and role-aware supervision, and later checkpoints tend to exhibit stronger physical and structural biases.
Motivated by Auto-Guidance~\cite{karras2024guidingdiffusionmodelbad} in image generation, VPT uses such an earlier intermediate model to guide the final model during inference.

Let \(\theta_i\) and \(\theta_f\) denote the intermediate model and the final model, respectively. 
We first apply classifier-free guidance~\cite{ho2022classifierfree} to each model:
\begin{equation}
\tilde{u}_{\theta}
=
u_{\theta}(Z, \emptyset, t_v, t_f, t_r)
+
s
\left(
u_{\theta}(Z, y, t_v, t_f, t_r)
-
u_{\theta}(Z, \emptyset, t_v, t_f, t_r)
\right),
\end{equation}
where \(Z=[z^v_{t_v}\oplus z^f_{t_a}\oplus z^r_{t_a}]\) and \(s\) is the text guidance scale, $t_v=t_f=t_r$.

The prediction of the final model is then extrapolated using the intermediate model as a reference:
\begin{equation}
\hat{u}
=
\tilde{u}_{\theta_f}
+
\gamma
\left(
\tilde{u}_{\theta_f}
-
\tilde{u}_{\theta_i}
\right),
\end{equation}
where \(\gamma\) controls the guidance strength. 
The intermediate model serves as a reference model, and the extrapolation amplifies the generation direction of the final model relative to the earlier model.

\vspace{-6pt}
\section{Experiment}
\vspace{-4pt}
\subsection{Implementation details}
\vspace{-4pt}
\textbf{Model setups.}
We implement VPT on Wan2.1-T2V-1.3B and Wan2.1-T2V-14B~\cite{wan2025wan}. 
Both models generate videos of 81 frames at $480 \times 832$ resolution and 16 FPS. 
VPT augments the RGB video input with two auxiliary modalities, optical flow and role maps, forming a 9-channel joint representation. 
Optical flow is extracted offline using RAFT~\cite{teed2020raft}. 
For role-map construction, we use Qwen3-VL-30B-A3B~\cite{bai2025qwen3vltechnicalreport} to assign physical roles to scene entities and SAM3~\cite{carion2026sam3segmentconcepts} for mask extraction. 
The role map contains four predefined categories: agent, controlled object, passive object, and background, and is replicated into three channels.
We expand the input and output projection layers to support the additional channels, and initialize the newly added parameters to zero. 
During fine-tuning, we train the expanded projection layers, timestamp embeddings, timestamp AdaLN layers, and LoRA \cite{hu2021loralowrankadaptationlarge} adapters in the attention blocks with LoRA rank and alpha both set to 32.

\textbf{Training details.}
We train VPT on the full WISA-80K dataset~\cite{wang2025wisa}. 
All experiments use AdamW~\cite{loshchilov2019decoupledweightdecayregularization} with a learning rate of $1 \times 10^{-5}$ for 2,000 optimization steps. 
The auxiliary loss weight is initialized as $\lambda_0=0.2$ and cosine-decayed to 0 throughout training. 
For Wan2.1-T2V-1.3B, fine-tuning is conducted on 8 NVIDIA A100 80GB GPUs with a per-GPU batch size of 6, resulting in a global batch size of 48. 
For Wan2.1-T2V-14B, fine-tuning is conducted on 32 NVIDIA A100 80GB GPUs with a per-GPU batch size of 1, resulting in a global batch size of 32.

\subsection{Evaluation Metrics}

We evaluate VPT from two aspects: physical consistency and general video quality.
VideoPhy~\cite{bansal2024videophy} measures physical plausibility in material interactions, including solid-solid, solid-fluid, and fluid-fluid scenarios, and reports Semantic Adherence (SA) and Physical Commonsense (PC).
VideoPhy-2~\cite{bansal2025videophy2} further evaluates more complex action-centric physical interactions with the same SA and PC metrics.
VBench~\cite{huang2023vbench} measures general video quality across quality, semantic, and overall scores.

\begin{table}[]
    \centering
    \caption{\textbf{Quantitative comparison on VideoPhy and VideoPhy-2 benchmark.} 
    We report Semantic Adherence (SA) and Physical Commonsense (PC) scores. 
    The best results are highlighted in \textbf{bold}.}
    \label{tab:videophy}
    \setlength{\tabcolsep}{4pt}
    \resizebox{\linewidth}{!}{%
        \begin{tabular}{lcccccccccc}
            \toprule
            \multirow{3}{*}[-2pt]{\textbf{\large Method}} &  \multicolumn{8}{c}{\textbf{VideoPhy}} & \multicolumn{2}{c}{\textbf{VideoPhy-2}} \\
            \cmidrule(lr){2-9} \cmidrule(lr){10-11}
            &
            \multicolumn{2}{c}{\textbf{Solid-Solid}} & 
            \multicolumn{2}{c}{\textbf{Solid-Fluid}} & 
            \multicolumn{2}{c}{\textbf{Fluid-Fluid}} & 
            \multicolumn{2}{c}{\textbf{Overall}} & \multicolumn{2}{c}{\textbf{Overall}} \\
            \cmidrule(lr){2-3} \cmidrule(lr){4-5} \cmidrule(lr){6-7} \cmidrule(lr){8-9} \cmidrule(lr){10-11}
            & \textbf{SA} & \textbf{PC} & \textbf{SA} & \textbf{PC} & \textbf{SA} & \textbf{PC} & \textbf{SA} & \textbf{PC} &  \textbf{SA} & \textbf{PC}\\
            \midrule
            Wan2.1-T2V-1.3B      & 51.1 & 22.3 & 45.2 & 19.1 & 45.4 & 23.6 & 47.7 & 21.2 & 19.3 & 53.7\\
            Wan2.1-T2V-1.3B (Full Fine-tune)      & 46.2 & 18.2 & 47.8 & 22.6 & 50.9 & 23.6 & 45.1 & 20.9 & 18.9 & 53.6\\
            Wan2.1-T2V-14B     &46.8&18.1 &65.7  &25.3  &54.5  &30.9 &56.1  &23.2 & 21.9 & 52.9 \\
            Wan2.1-T2V-14B (Full Fine-tune)   & 63.8   & 19.0 & 60.1 & 23.1 & 64.0 & 22.0 & 62.1  & 21.5 & 20.7 & 54.0\\
            Wan2.1-T2V-1.3B+VideoJAM & 45.4 & 13.2 & 56.1 & 27.3 & 40.1 & 30.9 & 49.1 & 22.1 & 20.6 & 54.0\\
            \midrule
            \textbf{Wan2.1-T2V-1.3B+VPT (Ours)} & \textbf{69.2} & 23.7 & 63.0 & 24.0 & \textbf{69.1} & 30.1 & 66.5 & 25.0 & 22.5 & 55.1\\
            \textbf{Wan2.1-T2V-14B+VPT (Ours)} & 67.8 & \textbf{27.3} & \textbf{67.8} & \textbf{31.5} & 67.3 & \textbf{32.7} & \textbf{67.7} & \textbf{30.0} & \textbf{23.3} & \textbf{59.9}\\
            \bottomrule
        \end{tabular}%
    }
\end{table}

\begin{table*}[t]
    \centering
    \scriptsize
    \caption{\textbf{Quantitative comparison on VBench.}
We use the official raw VBench prompts without prompt enhancement or official enhanced prompts, and evaluate all methods with 81 frames at \(480 \times 832\) resolution and 16 FPS.}
    \label{tab:vbench}
    \setlength{\tabcolsep}{1.5pt} 
    \resizebox{\linewidth}{!}{%
        \begin{tabular}{l ccccc ccccc c cc c}
            \toprule
            \multirow{2}{*}[-2ex]{\textbf{\large Method}} & 
            \multicolumn{5}{c}{\textbf{\small Temporal}} & 
            \multicolumn{5}{c}{\textbf{\small Semantic}} & 
            \multicolumn{1}{c}{\textbf{\small Spatial}} & 
            \multicolumn{2}{c}{\textbf{\small Summary}} &
            \multirow{2}{*}[-2ex]{\textbf{\small \shortstack{Overall\\Score}}} \\
            
            \cmidrule(lr){2-6} \cmidrule(lr){7-11} \cmidrule(lr){12-12} \cmidrule(lr){13-14}
            
            & \textbf{\shortstack{Subject\\Consistency}} 
            & \textbf{\shortstack{Background\\Consistency}} 
            & \textbf{\shortstack{Temporal\\Flickering}} 
            & \textbf{\shortstack{Motion\\Smoothness}} 
            & \textbf{\shortstack{Dynamic\\Degree}} 
            & \textbf{\shortstack{Object\\Class}} 
            & \raisebox{1.2ex}{\textbf{\shortstack Color}}
            & \textbf{\shortstack{Human\\Action}} 
            & \textbf{\shortstack{Multiple\\Objects}}
            & \raisebox{1.2ex}{\textbf{Scene}} 
            & \raisebox{1.2ex}{\textbf{Relationship}}
            & \textbf{\shortstack{Quality\\Score}}
            & \textbf{\shortstack{Semantic\\Score}}
            & \\ 
            \midrule 
            Wan2.1-T2V-1.3B & 91.83 & 94.71 & 99.17 & 96.51 & 65.00 & 76.09 & \textbf{89.93} & 74.60 & 53.66 & 20.03 & 62.37 & 79.81 & 65.43 & 76.93 \\
            Wan2.1-T2V-1.3B (Full Fine-tune) & 93.59 & 95.81 & \textbf{99.36} & 97.21 & 54.08 & 79.90 & 88.57 & 78.98 & 58.20 & \textbf{28.55} & 63.31 & 81.26 & 68.47 & 78.71 \\
            Wan2.1-T2V-1.3B+VideoJAM (Reimpl.) & 91.51 & 96.01 & 99.13 & 96.05 & \textbf{73.88} & \textbf{79.22} & 88.92 & \textbf{79.20} & \textbf{59.66} & 28.34 & 66.17 & 81.18 & \textbf{69.08} & 78.76 \\
            \midrule
            \textbf{\small Wan2.1-T2V-1.3B+VPT(Ours)} & \textbf{92.65} & \textbf{97.27} & 98.64 & \textbf{97.85} & 70.83 & 73.43 & 86.99 & 74.8 & 51.92 & 20.13 & \textbf{73.35} & \textbf{83.25} & 64.86 & \textbf{79.58} \\
            \bottomrule
        \end{tabular}%
    }
\end{table*}

\subsection{Quantitative Comparisons}
\label{sec:quantitative}
We conduct quantitative comparisons on VideoPhy, VideoPhy-2, and VBench in Tables~\ref{tab:videophy} and \ref{tab:vbench}. 

\paragraph{VideoPhy.}
Table~\ref{tab:videophy} shows that VPT consistently improves both SA and PC on VideoPhy. 
On Wan2.1-T2V-1.3B, VPT improves the overall SA from 47.7 to 66.5 and PC from 21.2 to 25.0, respectively. 
On Wan2.1-T2V-14B, VPT improves the overall SA from 56.1 to 67.7 and PC from 23.2 to 30.0, respectively. 
These results demonstrate that VPT is effective across different model scales. VPT also outperforms full fine-tuning and VideoJAM. 
Naive full fine-tuning does not consistently improve physical commonsense and can even reduce PC, indicating that simply adapting the whole model is insufficient for learning reliable physical priors. 
VideoJAM improves some categories but shows limited overall gains. 
In contrast, VPT improves both SA and PC across solid-solid, solid-fluid, and fluid-fluid interactions, demonstrating stronger and more robust physical modeling.

\paragraph{VideoPhy-2.}
Table~\ref{tab:videophy} reports results on VideoPhy-2, which contains more complex action-centric prompts and human-object interactions. 
With the 1.3B backbone, VPT improves SA from 19.3 to 22.5 and PC from 53.7 to 55.1. 
With the 14B backbone, VPT reaches 23.3 SA and 59.9 PC, obtaining the highest scores on both metrics. 
This verifies that VPT generalizes beyond simple material interactions and remains effective for dynamic scenarios involving contact, force, and object motion.

\paragraph{VBench.}
Table~\ref{tab:vbench} reports VBench results using the official raw prompts without prompt enhancement; all methods are sampled as 81 frames at \(480 \times 832\) resolution and 16 FPS.
VPT improves the overall score from 76.93 to 79.58 over Wan2.1-T2V-1.3B and achieves the highest quality score of 83.25.
It improves background consistency, motion smoothness, and spatial relationship, indicating more stable dynamics and better object relation modeling.

Overall, these quantitative results validate the effectiveness of VPT. 
By introducing role-aware joint representation, modality-decoupled denoising, and auxiliary loss annealing, VPT injects physically meaningful priors into pretrained video diffusion models through lightweight fine-tuning, improving physical consistency while preserving visual quality and prompt adherence.

\vspace{-4pt}
\subsection{Qualitative comparisons}
\begin{figure}
    \centering
    \includegraphics[width=\linewidth]{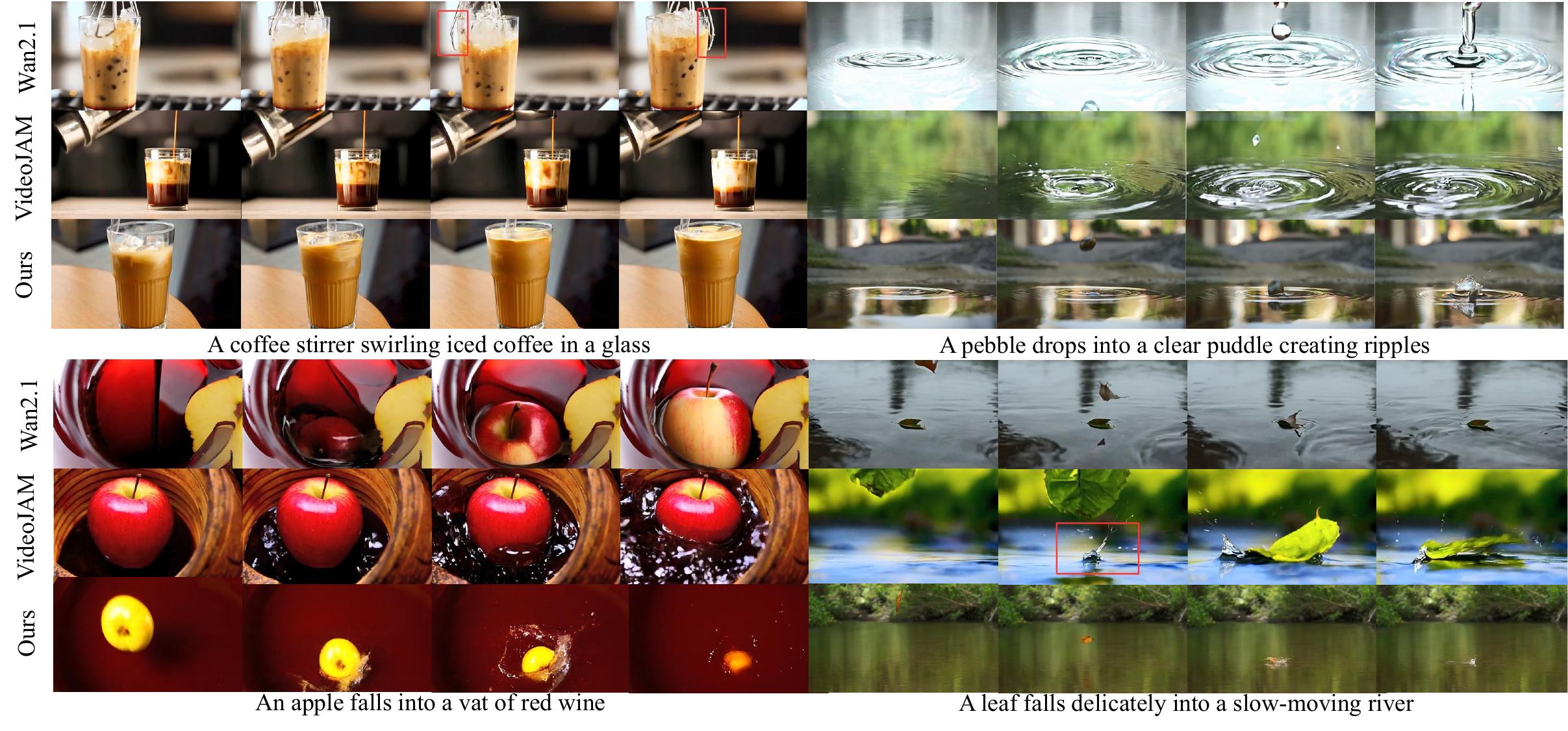}

\vspace{-5pt}
    \caption{
\textbf{Qualitative comparison on physically grounded video generation.}
Wan2.1-T2V-1.3B and VideoJAM (based on Wan2.1-T2V-1.3B) often suffer from physically implausible artifacts, such as unstable deformation, inconsistent ripples, and incoherent interaction dynamics.
VPT (based on Wan2.1-T2V-1.3B) better preserves object geometry and generates more realistic fluid responses and motion trajectories, demonstrating stronger physical consistency.
}
    \label{fig:compare}
\end{figure}

\vspace{-4pt}
We qualitatively compare VPT with Wan2.1-T2V~\cite{wan2025wan} and VideoJAM~\cite{chefer2025videojam} in Figure~\ref{fig:compare}, using prompts from Videophy~\cite{bansal2024videophy} benchmark. 
VPT consistently generates videos with higher visual quality, fewer motion distortions, and stronger physical consistency. 
In the coffee stirring case, our method produces coherent swirling motion following the stirrer trajectory, while the baselines show unstable liquid deformation. 
For the pebble-dropping example, VPT generates clearer impact dynamics and more plausible outward-propagating ripples, whereas the baselines exhibit weaker or spatially inconsistent water responses. 
In the apple-falling case, VPT better preserves the apple geometry and produces a realistic disturbance on the wine surface after impact. 
For the falling leaf, our method yields smoother motion and more natural interaction with the slow-moving river, avoiding abrupt deformation and unstable trajectories.
These results show that VPT effectively enhances physical consistency in video generation. 
With only lightweight fine-tuning, our method enables the pretrained video diffusion model to learn physically meaningful motion priors while preserving its visual generation capability. 
Additional qualitative comparisons and video results are provided in the supplementary material.
\subsection{Ablation Studies}
\label{sec:ablation}
\vspace{-4pt}

We conduct ablation studies to analyze the key components of VPT, including the training design, inference-time guidance, auxiliary physical modalities, CFG scale, and auxiliary-latent reconstruction quality. 
All ablations are conducted on Wan2.1-T2V-1.3B~\cite{wan2025wan} and evaluated on VideoPhy.

\textbf{Effect of training and inference strategy.}
Table~\ref{tab:ablation}(a) studies the effects of the proposed training strategy and inference-time guidance.
Naive joint training uses a constant loss weight and synchronized timesteps for all branches, disrupting the pretrained visual prior and causing severe video degeneration.
Its PC score is omitted because the generated videos collapse and fail the physical evaluation.
In contrast, our training strategy, which combines modality-decoupled denoising and auxiliary loss annealing, substantially improves the performance to 66.5 SA and 25.0 PC.
Applying Cross-step Auto-Guidance at inference further improves PC from 25.0 to 26.5 by using the 1,600-step model as the intermediate model with a guidance strength of 1.2, indicating stronger physical commonsense and motion dynamics while maintaining competitive semantic adherence.

\begin{table}[]
    \centering
    \caption{\textbf{Ablation studies on VPT.}
    We analyze the effects of training strategy,  inference-time guidance, auxiliary modalities, and auxiliary-latent reconstruction quality.
All ablations are evaluated on VideoPhy using Wan2.1-T2V-1.3B.
}
    \label{tab:ablation}
    \setlength{\tabcolsep}{3.pt}
    \vspace{1mm}
    \begin{minipage}[t]{0.37\linewidth}
        \centering
        \small
        \textbf{(a) Training and inference}\\[3pt]
        \begin{tabular}{lcc}
            \toprule
            \textbf{Setting} & \textbf{SA} & \textbf{PC} \\
            \midrule
            Naive Joint Training & 8.7 & - \\
            +Our Training Strategy & 66.5 & 25.0 \\
            +Cross-step Auto-Guidance & 64.5 & 26.5 \\
            \bottomrule
        \end{tabular}
    \end{minipage}
    \hfill
    \begin{minipage}[t]{0.3\linewidth}
        \centering
        \small
        \textbf{(b) Auxiliary modalities}\\[3pt]
        \begin{tabular}{lcc}
            \toprule
            \textbf{Setting} & \textbf{SA} & \textbf{PC} \\
            \midrule
            Baseline & 45.1 & 20.9 \\
            + Optical Flow & 49.3 & 22.4 \\
            + Role-aware Map & 66.5 & 25.0 \\
            \bottomrule
        \end{tabular}
    \end{minipage}
    \hfill
    \begin{minipage}[t]{0.3\linewidth}
        \centering
        \small
        \textbf{(c) VAE Reconstruction}\\[3pt]
        \begin{tabular}{lcc}
            \toprule
            \textbf{Modality} & \textbf{PSNR} & \textbf{SSIM} \\
            \midrule
            Optical Flow & 51.68 & 0.9951 \\
            Role-aware Map & 43.45 & 0.9928 \\
            \bottomrule
        \end{tabular}
    \end{minipage}
\end{table}

\begin{figure}
    \centering
    \includegraphics[width=\linewidth]{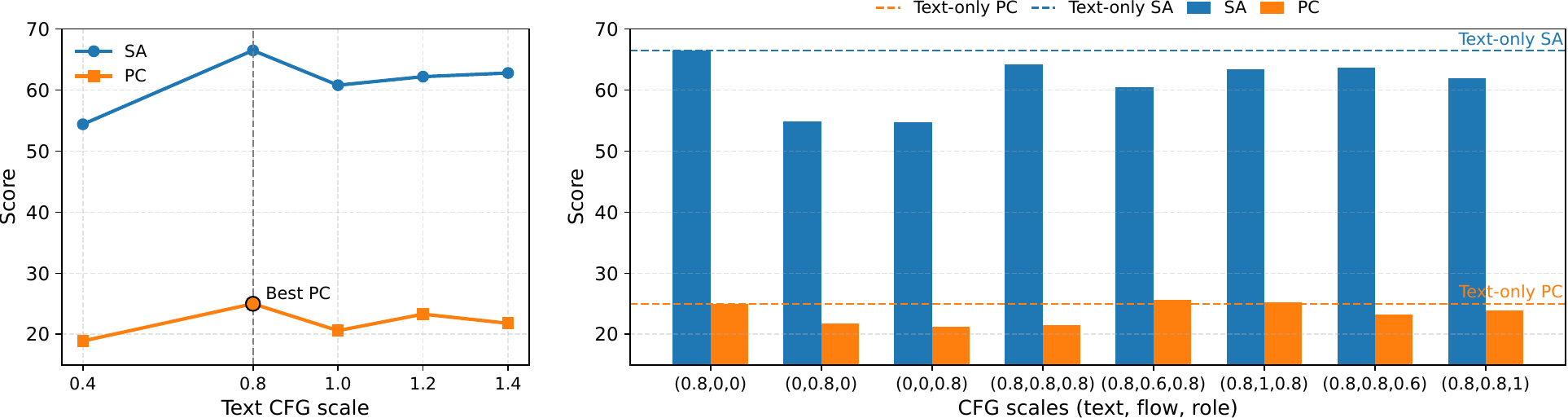}
    \caption{
\textbf{Ablation study on classifier-free guidance (CFG) scales.}
\textbf{Left}: text-only CFG sweep.
\textbf{Right}: CFG combinations over text, optical-flow, and role-aware latents.
Directly applying CFG to auxiliary modalities does not consistently outperform text-only guidance, as inference-time auxiliary predictions can contain larger errors and thus fail to provide reliable guidance.
}
\vspace{-1mm}
    \label{fig:cfg_sweep}
\end{figure}
\textbf{Effect of auxiliary modalities.}
Table~\ref{tab:ablation}(b) analyzes the contribution of different auxiliary modalities.
Starting from the baseline, adding optical flow improves SA from 45.1 to 49.3 and PC from 20.9 to 22.4, indicating that dense motion cues help the model learn more plausible temporal dynamics.
Further adding the role-aware map brings a larger improvement, increasing SA to 66.5 and PC to 25.0.
This suggests that role-aware maps provide complementary high-level physical structure, helping the model reason about active agents, controlled objects, passive objects, and background regions.

\textbf{Effect of guidance scale.}
We further study the effect of classifier-free guidance scales in Figure~\ref{fig:cfg_sweep}. 
The text-only CFG sweep shows that a moderate text guidance scale provides a good trade-off between semantic adherence and physical consistency. 
However, directly applying CFG to optical-flow or role-aware latents does not consistently improve performance over text-only guidance. 
This suggests that explicitly guided auxiliary predictions may introduce noisy inference-time signals, supporting our design choice of relying primarily on learned physical priors and Cross-step Auto-Guidance.

\textbf{Reconstruction quality of auxiliary modalities.}
Since VPT encodes auxiliary modalities into the latent space of the pretrained Wan video VAE, we evaluate its reconstruction quality on 1,000 randomly sampled training videos.
As shown in Table~\ref{tab:ablation}(c), both optical flow and role-aware maps achieve high PSNR and SSIM after VAE encoding and decoding, indicating that the shared video latent space preserves essential auxiliary physical cues for VPT supervision.

\section{Conclusion}

\vspace{-4pt}

We introduce VPT, a lightweight fine-tuning framework for improving physical consistency in text-to-video generation.
VPT combines optical flow and role-aware maps, modality-decoupled denoising with auxiliary loss annealing, and cross-step auto-guidance to inject physical motion priors while preserving visual quality.
Experiments on VideoPhy, VideoPhy-2, and VBench show consistent improvements in physical commonsense, semantic adherence, and generation quality, with qualitative results showing more coherent motion and realistic interactions.

\textbf{Limitations.}
VPT relies on automatically extracted optical flow, VLM-based role assignments, and segmentation masks, whose errors may affect physical-prior learning.
Our experiments are limited to Wan2.1 backbones and standard benchmarks, and due to GPU constraints, we have not evaluated broader video diffusion backbones or substantially larger physical-video datasets.

\bibliography{neurips_2026}
\bibliographystyle{abbrv}

\newpage

\appendix

\section{Comparison with Inner-Guidance}
\label{app:inner_guidance}

\paragraph{Inner-Guidance.}

Following VideoJAM~\cite{chefer2025videojam}, we adapt Inner-Guidance to VPT by applying separate guidance weights to text, optical-flow, and role-aware conditions.
Let \(u_{tfr}\) denote the velocity predicted with all conditions, i.e., text \(t\), optical flow \(f\), and role map \(r\).
We denote the predictions with one condition dropped as \(u_{\emptyset fr}\), \(u_{t\emptyset r}\), and \(u_{tf\emptyset}\), respectively.
Here, dropping the text condition replaces \(y\) with the null prompt \(\emptyset\), while dropping an auxiliary condition replaces the corresponding latent with zeros.
This is consistent with our training strategy, where the optical-flow and role-aware latents are independently dropped with probability 0.1 by setting them to zero.
Therefore, at inference time, the unconditional auxiliary condition is implemented by zeroing the corresponding latent at each denoising step; for example, \(u_{tf\emptyset}\) is obtained by setting the role-aware latent to zero while keeping the text and optical-flow conditions.
The guided visual prediction is computed as
\begin{equation}
\hat{u}_{\mathrm{IG}}
=
u_{tfr}
+
w_t
\left(
u_{tfr}
-
u_{\emptyset fr}
\right)
+
w_f
\left(
u_{tfr}
-
u_{t\emptyset r}
\right)
+
w_r
\left(
u_{tfr}
-
u_{tf\emptyset}
\right),
\label{eq:inner_guidance_vpt}
\end{equation}
where \(w_t\), \(w_f\), and \(w_r\) control the guidance strengths of text, optical flow, and role map, respectively.
During sampling, all modalities latents are initialized from noise and recursively predicted via inner-guidance, making the guidance sensitive to auxiliary prediction errors.

\paragraph{Comparison with Cross-step Auto-Guidance.}
Unlike inner-guidance, cross-step auto-guidance does not directly use recursively predicted auxiliary modalities as guidance.
Instead, it uses an intermediate model from the fine-tuning trajectory as a reference and amplifies the physical direction learned by the final model.
This strengthens physical dynamics while avoiding noisy auxiliary guidance.

\begin{table}[h]
    \centering
    \caption{\textbf{Comparison between Inner-Guidance and Cross-step Auto-Guidance.}
    We compare inference time and VideoPhy performance using Wan2.1-T2V-1.3B + VPT.
    Inner-Guidance applies separate guidance to text, optical-flow, and role-aware conditions, while Cross-step Auto-Guidance uses an intermediate model as reference guidance.
    }
    \label{tab:inner_guidance}
    \setlength{\tabcolsep}{5pt}
    \begin{tabular}{lccc}
        \toprule
        \textbf{Method} & \textbf{Time Cost} $\downarrow$ & \textbf{SA} $\uparrow$ & \textbf{PC} $\uparrow$ \\
        \midrule
        VPT w/o Text Guidance & 1m40s & 35.8 & 12.5 \\
        VPT w/ Text Guidance & 3m12s & \textbf{66.5} & 25.0 \\
        VPT + Inner-Guidance & 6m03s & 64.2 & 21.5 \\
        VPT + Cross-step Auto-Guidance & 6m19s & 64.5 & \textbf{26.5} \\
        \bottomrule
    \end{tabular}
\end{table}

Table~\ref{tab:inner_guidance} compares different inference guidance strategies.
Without text guidance, VPT has the lowest inference cost but much lower SA and PC.
Text guidance substantially improves both metrics, reaching 66.5 SA and 25.0 PC.
Applying inner-guidance requires additional forward passes for condition-dropped predictions and recursively relies on auxiliary latents, increasing the time cost to 6m03s.
However, due to noisy auxiliary predictions during inference, it reduces PC to 21.5 despite maintaining 64.2 SA.
This observation is consistent with the CFG sweep in Figure~\ref{fig:cfg_sweep}, where auxiliary-guided inner-guidance variants do not outperform the text-only guidance setting.
In contrast, cross-step auto-guidance achieves the best PC score of 26.5 with a comparable time cost, while maintaining competitive SA.
These results show that intermediate-model guidance provides a more effective trade-off than directly guiding generation with recursively predicted auxiliary modalities.

\section{Visualization of Auxiliary Reconstruction}
\label{app:aux_reconstruction}

Although the pretrained Wan video VAE is trained on RGB videos rather than optical-flow or role-aware maps, VPT encodes these auxiliary modalities into the same latent space for joint supervision.
To verify whether this encoding preserves the physical meaning of auxiliary signals, we provide qualitative reconstruction results in addition to the PSNR and SSIM results reported in Table~\ref{tab:ablation}(c).

\begin{figure}
    \centering
    \includegraphics[width=\linewidth]{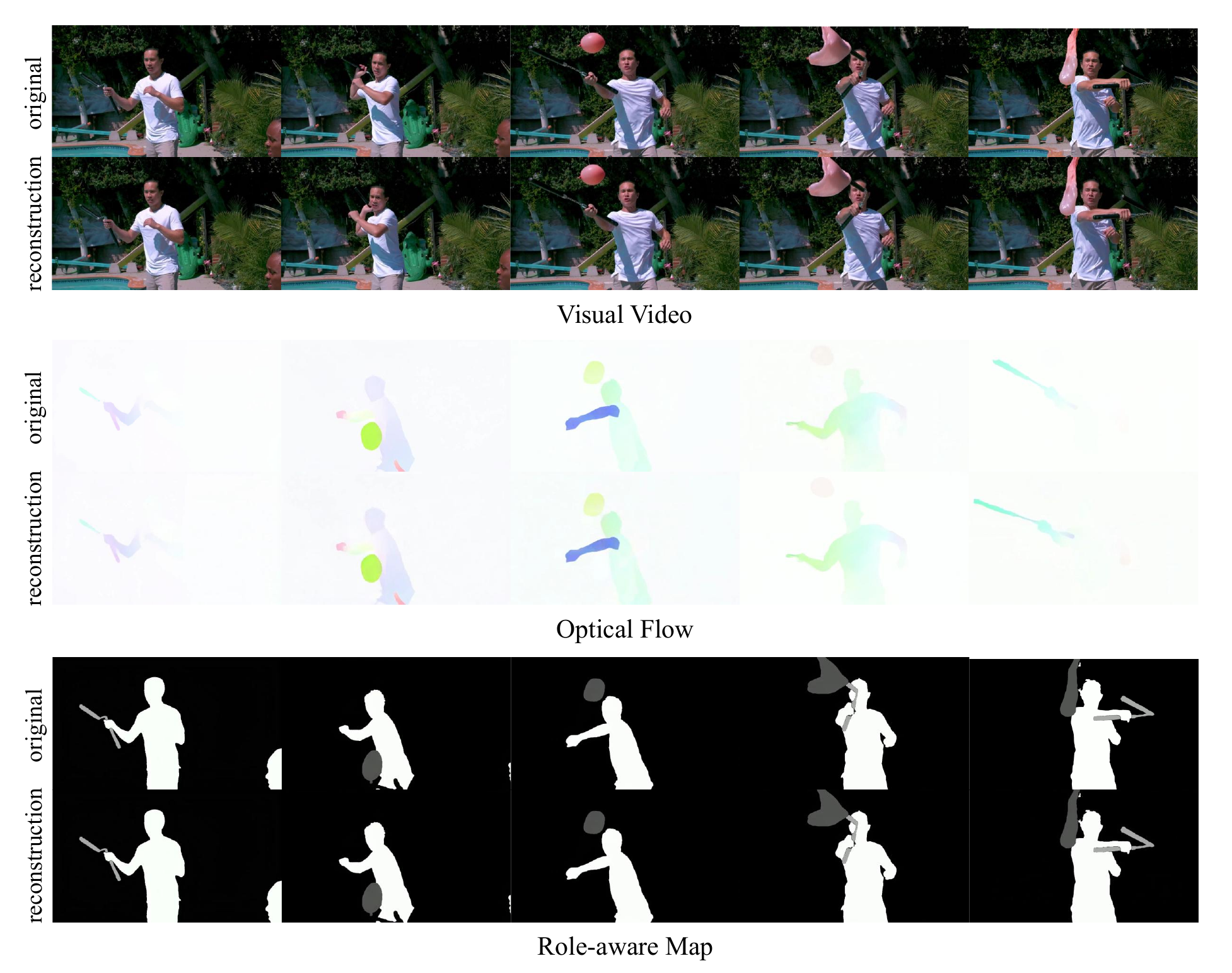}
    \caption{
\textbf{VAE reconstruction of RGB video, optical flow, and role-aware maps.}
For each modality, we show the original input and its VAE reconstruction.
The reconstructed optical flow retains the original motion direction and temporal trajectory, and the reconstructed role-aware map preserves entity regions and role labels.
}

    \label{fig:aux_recon}
\end{figure}

As shown in Figure~\ref{fig:aux_recon}, the decoded optical-flow videos closely match the original optical-flow inputs after VAE encoding and decoding.
The reconstructed flow preserves the main motion direction, spatial displacement pattern, and temporal evolution across frames, rather than merely producing visually similar appearance.
Similarly, the reconstructed role-aware maps preserve the entity regions, role labels, and object boundaries, indicating that the VAE latent space retains the essential structure needed for role-aware physical supervision.
Visual comparisons of the original and reconstructed videos, optical-flow videos, and role-map videos are included in the supplementary folder \texttt{./reconstruction\_examples}.

These visualizations suggest that the shared video VAE latent space can faithfully represent the auxiliary modalities used in VPT.
Together with the quantitative reconstruction metrics in Table~\ref{tab:ablation}(c), they support the feasibility of applying auxiliary supervision in the pretrained video latent space.

\section{More qualitative results}

\begin{figure}
    \centering
    \includegraphics[width=\linewidth]{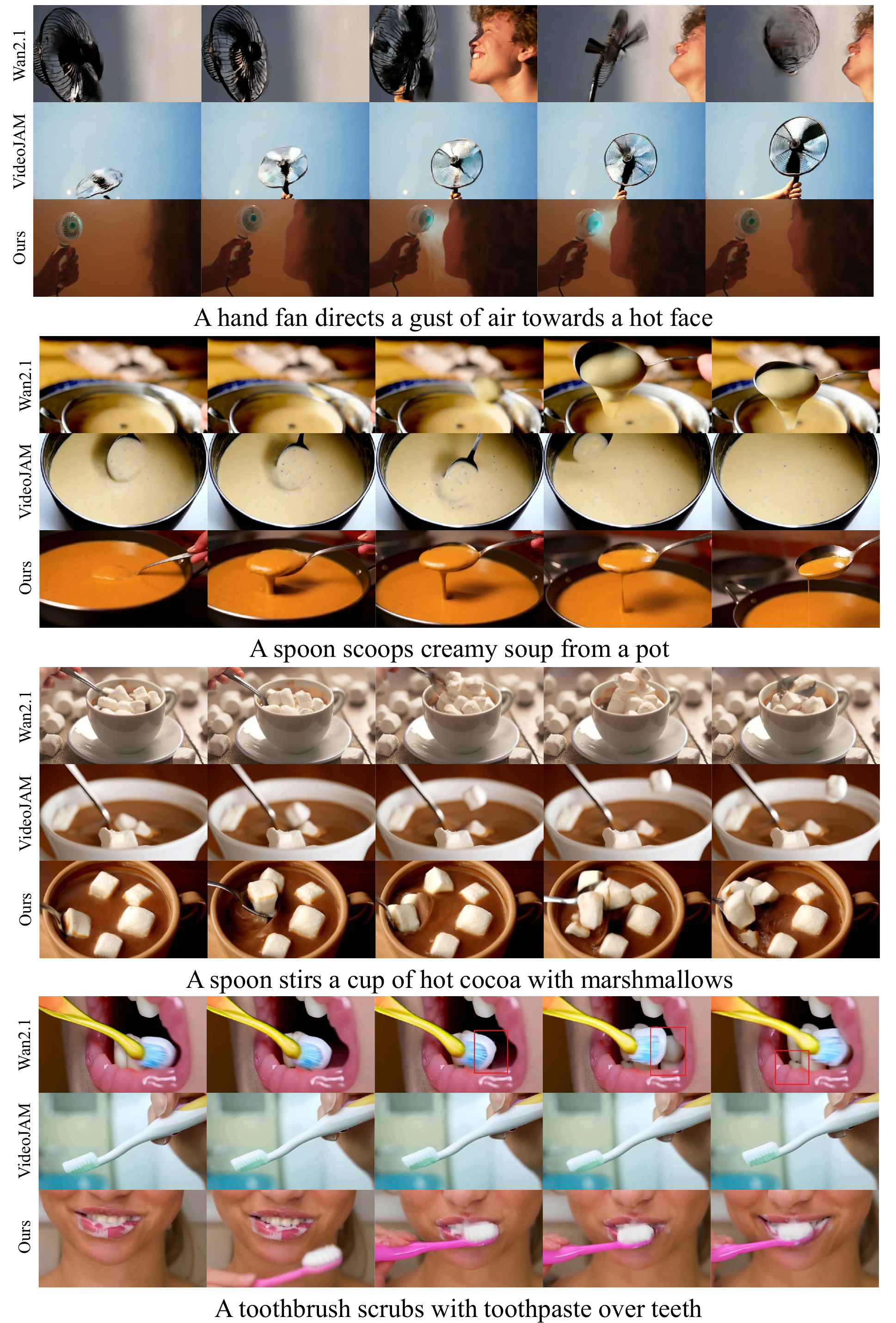}
    \caption{Qualitative results. The first row displays the outcomes of Wan2.1-1.3B, the second row shows the results of VideoJAM, and the third row presents the results of our VPT.}
    \label{fig:appendix1}
\end{figure}

\begin{figure}
    \centering
    \includegraphics[width=\linewidth]{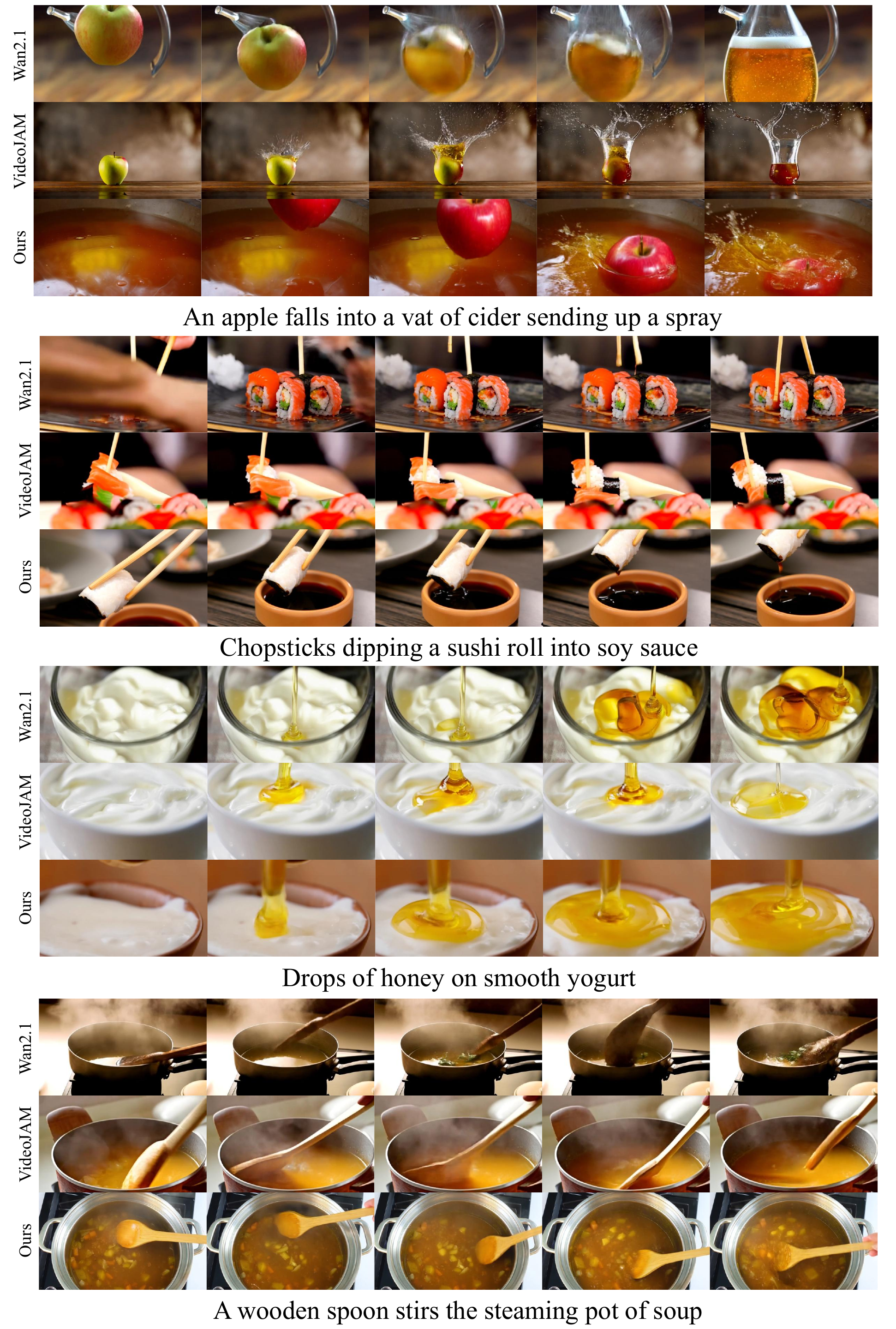}
    \caption{Qualitative results. The first row displays the outcomes of Wan2.1-1.3B, the second row shows the results of VideoJAM, and the third row presents the results of our VPT.}
    \label{fig:appendix2}
\end{figure}

In this section, we present additional qualitative comparisons between our method and baselines.
For each prompt, we show representative frames from generated videos.
Red rectangles highlight regions where physical commonsense is violated.
As shown in Figure~\ref{fig:appendix1} and Figure~\ref{fig:appendix2}, our method exhibits stronger object motion dynamics and better physical consistency.
Additional video comparisons are included in the supplementary folder \texttt{./videos}.
For convenient viewing, we also provide a local comparison webpage, \texttt{./Videos\_Comparison.html}, which can be directly opened in a browser.

\newpage
\section*{NeurIPS Paper Checklist}

\begin{enumerate}

\item {\bf Claims}
    \item[] Question: Do the main claims made in the abstract and introduction accurately reflect the paper's contributions and scope?
    \item[] Answer: \answerYes{} 
    \item[] Justification: The abstract and introduction accurately state the main contributions and scope of the paper: VPT improves physical consistency in video diffusion models through role-aware joint representation, modality-decoupled denoising, auxiliary loss annealing, and cross-step auto-guidance. The empirical claims are supported by quantitative and qualitative results on VideoPhy, VideoPhy-2, and VBench.
    \item[] Guidelines:
    \begin{itemize}
        \item The answer \answerNA{} means that the abstract and introduction do not include the claims made in the paper.
        \item The abstract and/or introduction should clearly state the claims made, including the contributions made in the paper and important assumptions and limitations. A \answerNo{} or \answerNA{} answer to this question will not be perceived well by the reviewers. 
        \item The claims made should match theoretical and experimental results, and reflect how much the results can be expected to generalize to other settings. 
        \item It is fine to include aspirational goals as motivation as long as it is clear that these goals are not attained by the paper. 
    \end{itemize}

\item {\bf Limitations}
    \item[] Question: Does the paper discuss the limitations of the work performed by the authors?
    \item[] Answer: \answerYes{} 
    \item[] Justification: VPT depends on the quality of automatically extracted optical flow, VLM-based role assignment, and segmentation masks. Our experiments are limited to Wan2.1 backbones and the VideoPhy, VideoPhy-2, and VBench benchmarks. Due to limited GPU resources, we did not evaluate more video diffusion backbones, and because existing physical-video datasets are still limited in scale, we have not explored fine-tuning on substantially larger physical datasets.
    \item[] Guidelines:
    \begin{itemize}
        \item The answer \answerNA{} means that the paper has no limitation while the answer \answerNo{} means that the paper has limitations, but those are not discussed in the paper. 
        \item The authors are encouraged to create a separate ``Limitations'' section in their paper.
        \item The paper should point out any strong assumptions and how robust the results are to violations of these assumptions (e.g., independence assumptions, noiseless settings, model well-specification, asymptotic approximations only holding locally). The authors should reflect on how these assumptions might be violated in practice and what the implications would be.
        \item The authors should reflect on the scope of the claims made, e.g., if the approach was only tested on a few datasets or with a few runs. In general, empirical results often depend on implicit assumptions, which should be articulated.
        \item The authors should reflect on the factors that influence the performance of the approach. For example, a facial recognition algorithm may perform poorly when image resolution is low or images are taken in low lighting. Or a speech-to-text system might not be used reliably to provide closed captions for online lectures because it fails to handle technical jargon.
        \item The authors should discuss the computational efficiency of the proposed algorithms and how they scale with dataset size.
        \item If applicable, the authors should discuss possible limitations of their approach to address problems of privacy and fairness.
        \item While the authors might fear that complete honesty about limitations might be used by reviewers as grounds for rejection, a worse outcome might be that reviewers discover limitations that aren't acknowledged in the paper. The authors should use their best judgment and recognize that individual actions in favor of transparency play an important role in developing norms that preserve the integrity of the community. Reviewers will be specifically instructed to not penalize honesty concerning limitations.
    \end{itemize}

\item {\bf Theory assumptions and proofs}
    \item[] Question: For each theoretical result, does the paper provide the full set of assumptions and a complete (and correct) proof?
    \item[] Answer: \answerNA{} 
    \item[] Justification: The paper does not present formal theoretical results, theorems, or proofs. The mathematical formulations are used to define the training objectives and inference guidance strategy.
    \item[] Guidelines:
    \begin{itemize}
        \item The answer \answerNA{} means that the paper does not include theoretical results. 
        \item All the theorems, formulas, and proofs in the paper should be numbered and cross-referenced.
        \item All assumptions should be clearly stated or referenced in the statement of any theorems.
        \item The proofs can either appear in the main paper or the supplemental material, but if they appear in the supplemental material, the authors are encouraged to provide a short proof sketch to provide intuition. 
        \item Inversely, any informal proof provided in the core of the paper should be complemented by formal proofs provided in appendix or supplemental material.
        \item Theorems and Lemmas that the proof relies upon should be properly referenced. 
    \end{itemize}

    \item {\bf Experimental result reproducibility}
    \item[] Question: Does the paper fully disclose all the information needed to reproduce the main experimental results of the paper to the extent that it affects the main claims and/or conclusions of the paper (regardless of whether the code and data are provided or not)?
    \item[] Answer: \answerYes{} 
    \item[] Justification: Sections 3 and 4 describe the model modifications, auxiliary modality construction, training objective, evaluation benchmarks, optimizer, learning rate, training steps, batch size, and compute setup. We will also release code, checkpoints, and role-map processing scripts to enable reproduction of the main experimental results.
    \item[] Guidelines:
    \begin{itemize}
        \item The answer \answerNA{} means that the paper does not include experiments.
        \item If the paper includes experiments, a \answerNo{} answer to this question will not be perceived well by the reviewers: Making the paper reproducible is important, regardless of whether the code and data are provided or not.
        \item If the contribution is a dataset and\slash or model, the authors should describe the steps taken to make their results reproducible or verifiable. 
        \item Depending on the contribution, reproducibility can be accomplished in various ways. For example, if the contribution is a novel architecture, describing the architecture fully might suffice, or if the contribution is a specific model and empirical evaluation, it may be necessary to either make it possible for others to replicate the model with the same dataset, or provide access to the model. In general. releasing code and data is often one good way to accomplish this, but reproducibility can also be provided via detailed instructions for how to replicate the results, access to a hosted model (e.g., in the case of a large language model), releasing of a model checkpoint, or other means that are appropriate to the research performed.
        \item While NeurIPS does not require releasing code, the conference does require all submissions to provide some reasonable avenue for reproducibility, which may depend on the nature of the contribution. For example
        \begin{enumerate}
            \item If the contribution is primarily a new algorithm, the paper should make it clear how to reproduce that algorithm.
            \item If the contribution is primarily a new model architecture, the paper should describe the architecture clearly and fully.
            \item If the contribution is a new model (e.g., a large language model), then there should either be a way to access this model for reproducing the results or a way to reproduce the model (e.g., with an open-source dataset or instructions for how to construct the dataset).
            \item We recognize that reproducibility may be tricky in some cases, in which case authors are welcome to describe the particular way they provide for reproducibility. In the case of closed-source models, it may be that access to the model is limited in some way (e.g., to registered users), but it should be possible for other researchers to have some path to reproducing or verifying the results.
        \end{enumerate}
    \end{itemize}

\item {\bf Open access to data and code}
    \item[] Question: Does the paper provide open access to the data and code, with sufficient instructions to faithfully reproduce the main experimental results, as described in supplemental material?
    \item[] Answer: \answerYes{} 
    \item[] Justification: We will provide open-source code, fine-tuned checkpoints, role-map processing scripts, and instructions for data preparation and evaluation. The release will include sufficient instructions to reproduce the proposed method and the main benchmark evaluations.
    \item[] Guidelines:
    \begin{itemize}
        \item The answer \answerNA{} means that paper does not include experiments requiring code.
        \item Please see the NeurIPS code and data submission guidelines (\url{https://neurips.cc/public/guides/CodeSubmissionPolicy}) for more details.
        \item While we encourage the release of code and data, we understand that this might not be possible, so \answerNo{} is an acceptable answer. Papers cannot be rejected simply for not including code, unless this is central to the contribution (e.g., for a new open-source benchmark).
        \item The instructions should contain the exact command and environment needed to run to reproduce the results. See the NeurIPS code and data submission guidelines (\url{https://neurips.cc/public/guides/CodeSubmissionPolicy}) for more details.
        \item The authors should provide instructions on data access and preparation, including how to access the raw data, preprocessed data, intermediate data, and generated data, etc.
        \item The authors should provide scripts to reproduce all experimental results for the new proposed method and baselines. If only a subset of experiments are reproducible, they should state which ones are omitted from the script and why.
        \item At submission time, to preserve anonymity, the authors should release anonymized versions (if applicable).
        \item Providing as much information as possible in supplemental material (appended to the paper) is recommended, but including URLs to data and code is permitted.
    \end{itemize}

\item {\bf Experimental setting/details}
    \item[] Question: Does the paper specify all the training and test details (e.g., data splits, hyperparameters, how they were chosen, type of optimizer) necessary to understand the results?
    \item[] Answer: \answerYes{} 
    \item[] Justification: The paper specifies the backbone models, video resolution and frame rate, auxiliary modalities, role-map construction procedure, trainable parameters, optimizer, learning rate, number of training steps, auxiliary loss schedule, batch size, and evaluation metrics.
    \item[] Guidelines:
    \begin{itemize}
        \item The answer \answerNA{} means that the paper does not include experiments.
        \item The experimental setting should be presented in the core of the paper to a level of detail that is necessary to appreciate the results and make sense of them.
        \item The full details can be provided either with the code, in appendix, or as supplemental material.
    \end{itemize}

\item {\bf Experiment statistical significance}
    \item[] Question: Does the paper report error bars suitably and correctly defined or other appropriate information about the statistical significance of the experiments?
    \item[] Answer: \answerNo{} 
    \item[] Justification: The paper reports point estimates on official automatic evaluators but does not report error bars, confidence intervals, statistical significance tests, or repeated-run variance. This is mainly due to the computational cost of repeatedly fine-tuning and evaluating large video diffusion models; we follow the official benchmark protocols and report the standard metrics.
    \item[] Guidelines:
    \begin{itemize}
        \item The answer \answerNA{} means that the paper does not include experiments.
        \item The authors should answer \answerYes{} if the results are accompanied by error bars, confidence intervals, or statistical significance tests, at least for the experiments that support the main claims of the paper.
        \item The factors of variability that the error bars are capturing should be clearly stated (for example, train/test split, initialization, random drawing of some parameter, or overall run with given experimental conditions).
        \item The method for calculating the error bars should be explained (closed form formula, call to a library function, bootstrap, etc.)
        \item The assumptions made should be given (e.g., Normally distributed errors).
        \item It should be clear whether the error bar is the standard deviation or the standard error of the mean.
        \item It is OK to report 1-sigma error bars, but one should state it. The authors should preferably report a 2-sigma error bar than state that they have a 96\% CI, if the hypothesis of Normality of errors is not verified.
        \item For asymmetric distributions, the authors should be careful not to show in tables or figures symmetric error bars that would yield results that are out of range (e.g., negative error rates).
        \item If error bars are reported in tables or plots, the authors should explain in the text how they were calculated and reference the corresponding figures or tables in the text.
    \end{itemize}

\item {\bf Experiments compute resources}
    \item[] Question: For each experiment, does the paper provide sufficient information on the computer resources (type of compute workers, memory, time of execution) needed to reproduce the experiments?
    \item[] Answer: \answerYes{} 
    \item[] Justification: We report the compute resources used to train our VPT models. For Wan2.1-T2V-1.3B, VPT is trained for 2,000 optimization steps on 8 NVIDIA A100 80GB GPUs, with a per-GPU batch size of 6 and a global batch size of 48. The training takes approximately 4 days.
    \item[] Guidelines:
    \begin{itemize}
        \item The answer \answerNA{} means that the paper does not include experiments.
        \item The paper should indicate the type of compute workers CPU or GPU, internal cluster, or cloud provider, including relevant memory and storage.
        \item The paper should provide the amount of compute required for each of the individual experimental runs as well as estimate the total compute. 
        \item The paper should disclose whether the full research project required more compute than the experiments reported in the paper (e.g., preliminary or failed experiments that didn't make it into the paper). 
    \end{itemize}
    
\item {\bf Code of ethics}
    \item[] Question: Does the research conducted in the paper conform, in every respect, with the NeurIPS Code of Ethics \url{https://neurips.cc/public/EthicsGuidelines}?
    \item[] Answer: \answerYes{} 
    \item[] Justification: The research conforms to the NeurIPS Code of Ethics. It uses existing models, datasets, and automatic evaluation benchmarks, and does not involve human-subject experiments or private personal data collection.
    \item[] Guidelines:
    \begin{itemize}
        \item The answer \answerNA{} means that the authors have not reviewed the NeurIPS Code of Ethics.
        \item If the authors answer \answerNo, they should explain the special circumstances that require a deviation from the Code of Ethics.
        \item The authors should make sure to preserve anonymity (e.g., if there is a special consideration due to laws or regulations in their jurisdiction).
    \end{itemize}

\item {\bf Broader impacts}
    \item[] Question: Does the paper discuss both potential positive societal impacts and negative societal impacts of the work performed?
    \item[] Answer: \answerYes{} 
    \item[] Justification: We discuss broader impacts in the Broader Impact section. Potential positive impacts include improving physically grounded video simulation, education, creative tools, and embodied AI research; potential negative impacts include misuse of more realistic video generation for deceptive synthetic media, and we discuss responsible release and usage restrictions as mitigations.
    \item[] Guidelines:
    \begin{itemize}
        \item The answer \answerNA{} means that there is no societal impact of the work performed.
        \item If the authors answer \answerNA{} or \answerNo, they should explain why their work has no societal impact or why the paper does not address societal impact.
        \item Examples of negative societal impacts include potential malicious or unintended uses (e.g., disinformation, generating fake profiles, surveillance), fairness considerations (e.g., deployment of technologies that could make decisions that unfairly impact specific groups), privacy considerations, and security considerations.
        \item The conference expects that many papers will be foundational research and not tied to particular applications, let alone deployments. However, if there is a direct path to any negative applications, the authors should point it out. For example, it is legitimate to point out that an improvement in the quality of generative models could be used to generate Deepfakes for disinformation. On the other hand, it is not needed to point out that a generic algorithm for optimizing neural networks could enable people to train models that generate Deepfakes faster.
        \item The authors should consider possible harms that could arise when the technology is being used as intended and functioning correctly, harms that could arise when the technology is being used as intended but gives incorrect results, and harms following from (intentional or unintentional) misuse of the technology.
        \item If there are negative societal impacts, the authors could also discuss possible mitigation strategies (e.g., gated release of models, providing defenses in addition to attacks, mechanisms for monitoring misuse, mechanisms to monitor how a system learns from feedback over time, improving the efficiency and accessibility of ML).
    \end{itemize}
    
\item {\bf Safeguards}
    \item[] Question: Does the paper describe safeguards that have been put in place for responsible release of data or models that have a high risk for misuse (e.g., pre-trained language models, image generators, or scraped datasets)?
    \item[] Answer: \answerNA{} 
    \item[] Justification: This work does not release a new high-risk generative model, a new dataset, or any scraped dataset. The released asset is the code for the VPT training framework, i.e., a fine-tuning framework that implements the methods described in the paper, together with data-processing scripts and experimental hyperparameters. The backbone models used in this work, such as Wan2.1-T2V, are already publicly available, and the training and evaluation assets, including WISA, VideoPhy, and VBench, are existing publicly released resources. Therefore, no additional safeguards for releasing high-risk models or datasets are required.
    \item[] Guidelines:
    \begin{itemize}
        \item The answer \answerNA{} means that the paper poses no such risks.
        \item Released models that have a high risk for misuse or dual-use should be released with necessary safeguards to allow for controlled use of the model, for example by requiring that users adhere to usage guidelines or restrictions to access the model or implementing safety filters. 
        \item Datasets that have been scraped from the Internet could pose safety risks. The authors should describe how they avoided releasing unsafe images.
        \item We recognize that providing effective safeguards is challenging, and many papers do not require this, but we encourage authors to take this into account and make a best faith effort.
    \end{itemize}

\item {\bf Licenses for existing assets}
    \item[] Question: Are the creators or original owners of assets (e.g., code, data, models), used in the paper, properly credited and are the license and terms of use explicitly mentioned and properly respected?
    \item[] Answer: \answerYes{} 
    \item[] Justification: We properly credit the original creators of all existing assets used in this work, including the datasets, pretrained models, evaluation benchmarks, and auxiliary tools, by citing their original papers. The WISA dataset is derived from publicly available/public-domain sources, and Wan2.1-T2V is an open-source model released by its authors. All pretrained models and tools used in this work are publicly released under their respective licenses, and our use of these assets is limited to research purposes permitted by their licenses and terms of use.
    \item[] Guidelines:
    \begin{itemize}
        \item The answer \answerNA{} means that the paper does not use existing assets.
        \item The authors should cite the original paper that produced the code package or dataset.
        \item The authors should state which version of the asset is used and, if possible, include a URL.
        \item The name of the license (e.g., CC-BY 4.0) should be included for each asset.
        \item For scraped data from a particular source (e.g., website), the copyright and terms of service of that source should be provided.
        \item If assets are released, the license, copyright information, and terms of use in the package should be provided. For popular datasets, \url{paperswithcode.com/datasets} has curated licenses for some datasets. Their licensing guide can help determine the license of a dataset.
        \item For existing datasets that are re-packaged, both the original license and the license of the derived asset (if it has changed) should be provided.
        \item If this information is not available online, the authors are encouraged to reach out to the asset's creators.
    \end{itemize}

\item {\bf New assets}
    \item[] Question: Are new assets introduced in the paper well documented and is the documentation provided alongside the assets?
    \item[] Answer: \answerYes{} 
    \item[] Justification: The main new asset introduced in this work is the codebase for the VPT training framework, including the implementation of role-aware joint training, modality-decoupled denoising, auxiliary loss annealing, and cross-step auto-guidance, as well as data-processing scripts and complete experimental hyperparameters. The code will be released anonymously during submission to preserve double-blind review; upon acceptance, the repository will be de-anonymized and publicly released.
    \item[] Guidelines:
    \begin{itemize}
        \item The answer \answerNA{} means that the paper does not release new assets.
        \item Researchers should communicate the details of the dataset\slash code\slash model as part of their submissions via structured templates. This includes details about training, license, limitations, etc. 
        \item The paper should discuss whether and how consent was obtained from people whose asset is used.
        \item At submission time, remember to anonymize your assets (if applicable). You can either create an anonymized URL or include an anonymized zip file.
    \end{itemize}

\item {\bf Crowdsourcing and research with human subjects}
    \item[] Question: For crowdsourcing experiments and research with human subjects, does the paper include the full text of instructions given to participants and screenshots, if applicable, as well as details about compensation (if any)? 
    \item[] Answer: \answerNA{} 
    \item[] Justification: The paper does not involve crowdsourcing experiments or research with human subjects. The evaluations are conducted using automatic benchmark evaluators.
    \item[] Guidelines:
    \begin{itemize}
        \item The answer \answerNA{} means that the paper does not involve crowdsourcing nor research with human subjects.
        \item Including this information in the supplemental material is fine, but if the main contribution of the paper involves human subjects, then as much detail as possible should be included in the main paper. 
        \item According to the NeurIPS Code of Ethics, workers involved in data collection, curation, or other labor should be paid at least the minimum wage in the country of the data collector. 
    \end{itemize}

\item {\bf Institutional review board (IRB) approvals or equivalent for research with human subjects}
    \item[] Question: Does the paper describe potential risks incurred by study participants, whether such risks were disclosed to the subjects, and whether Institutional Review Board (IRB) approvals (or an equivalent approval/review based on the requirements of your country or institution) were obtained?
    \item[] Answer: \answerNA{} 
    \item[] Justification: The paper does not involve crowdsourcing or human-subject research, so IRB approval or equivalent review is not applicable.
    \item[] Guidelines:
    \begin{itemize}
        \item The answer \answerNA{} means that the paper does not involve crowdsourcing nor research with human subjects.
        \item Depending on the country in which research is conducted, IRB approval (or equivalent) may be required for any human subjects research. If you obtained IRB approval, you should clearly state this in the paper. 
        \item We recognize that the procedures for this may vary significantly between institutions and locations, and we expect authors to adhere to the NeurIPS Code of Ethics and the guidelines for their institution. 
        \item For initial submissions, do not include any information that would break anonymity (if applicable), such as the institution conducting the review.
    \end{itemize}

\item {\bf Declaration of LLM usage}
    \item[] Question: Does the paper describe the usage of LLMs if it is an important, original, or non-standard component of the core methods in this research? Note that if the LLM is used only for writing, editing, or formatting purposes and does \emph{not} impact the core methodology, scientific rigor, or originality of the research, declaration is not required.
    \item[] Answer: \answerYes{} 
    \item[] Justification: The method uses Qwen3-VL as an important component for assigning physical roles to scene entities during role-map construction. This usage directly affects the proposed role-aware representation and is described in the methodology and implementation details.
    \item[] Guidelines:
    \begin{itemize}
        \item The answer \answerNA{} means that the core method development in this research does not involve LLMs as any important, original, or non-standard components.
        \item Please refer to our LLM policy in the NeurIPS handbook for what should or should not be described.
    \end{itemize}

\end{enumerate}

\end{document}